\documentclass[conference]{IEEEtran}
\IEEEoverridecommandlockouts
\usepackage{cite}
\usepackage{amsmath,amssymb,amsfonts}
\usepackage{algorithmic}
\usepackage{graphicx}
\usepackage{textcomp}
\usepackage{xcolor}
\usepackage[utf8]{inputenc}
\usepackage{textcomp}
\usepackage{multirow}
\def\BibTeX{{\rm B\kern-.05em{\sc i\kern-.025em b}\kern-.08em
    T\kern-.1667em\lower.7ex\hbox{E}\kern-.125emX}}
\begin{document}

\title{Quantized Machine Learning Models for Medical Imaging in Low-Resource Healthcare Settings\\
\thanks{Identify applicable funding agency here. If none, delete this.}
}

\author{
\IEEEauthorblockN{Sumanth Meenan Kanneti}
\IEEEauthorblockA{\textit{Georgia State University} \\
Atlanta, USA \\
sumanthmeenan.kanneti123@gmail.com}

\and
\IEEEauthorblockN{Aryan Shah}
\textit{University of Southern California}\\
Los Angeles, USA \\
aryanutp@usc.edu
}

\maketitle

 \begin{abstract}
Deep learning models have shown strong performance in medical image analysis, but deploying them in low-resource clinical environments remains difficult due to computational, memory, and power constraints. This paper presents a multi-strategy compression framework for brain tumor classification from MRI, encompassing quantization-aware training, knowledge distillation from a DenseNet-101 teacher to a compact DenseNet-32 student with low-bit post-training quantization, and Float16 post-training quantization on a lightweight MobileNetV2 backbone. Using a multi-class brain tumor MRI dataset containing glioma, meningioma, pituitary tumors, and healthy controls, we provide full experimental validation of the MobileNetV2-based pipeline, training the classifier through a three-stage transfer learning process and applying Float16 quantization via TensorFlow Lite. The DenseNet-based distillation and quantization-aware training strategies are described as complementary compression approaches within the framework, with their complete empirical evaluation reserved for future work. Experimental results on the MobileNetV2 pipeline show that the quantized model achieves 82.37\% validation accuracy compared to the 82.20\% full-precision baseline, reducing model size from 35.34 MB to 5.76 MB, a 6.14×$\times$
compression ratio with no meaningful accuracy loss. Per-class evaluation confirms that quantization preserves diagnostic performance uniformly across all four tumor categories. These findings demonstrate that lightweight quantized models can deliver clinically viable brain tumor screening in resource-constrained healthcare settings.

\end{abstract}

\begin{IEEEkeywords} 
Abstract, Introduction,, ,, ,, , ,References\end{IEEEkeywords}
\section{Introduction}
Deep learning has rapidly transformed medical image analysis by enabling automated detection and classification of complex pathological patterns with accuracy approaching that of expert clinicians. Convolutional Neural Networks (CNNs), in particular, have demonstrated strong performance across a wide range of radiological tasks, including tumor detection, organ segmentation, and disease staging from MRI, CT, and X-ray images. In neuro-oncology, MRI-based brain tumor classification has benefited significantly from deep architectures such as DenseNet, ResNet, and MobileNet, which leverage deep feature reuse, residual connections, and efficient inverted bottleneck designs to capture subtle structural variations across tumor types. These advances promise faster screening, reduced diagnostic burden, and improved clinical decision support. However, most high-performing deep models are computationally intensive, requiring GPUs, large memory footprints, and stable power supply, assumptions that do not hold in many real-world healthcare environments. Rural clinics, mobile diagnostic units, and under-resourced hospitals often operate with limited hardware, low-power CPUs, or edge devices where deploying full-precision deep networks is impractical. As a result, there is a growing gap between algorithmic capability and deployable clinical utility. Bridging this gap requires not only accurate models but also efficient ones, models that preserve diagnostic performance while operating under strict compute and memory constraints.

Model compression and numerical efficiency techniques have therefore become central to the emerging field of edge medical AI. Among these, quantization, the reduction of numerical precision of model weights and activations, has emerged as one of the most practical and hardware-compatible approaches for accelerating inference and reducing model size. By converting floating-point parameters (typically 32-bit) into lower-bit representations such as Float16 or INT8, quantization can substantially decrease memory usage and increase throughput on commodity processors and specialized accelerators. Two major paradigms exist: post-training quantization (PTQ), which compresses a trained model after convergence, and quantization-aware training (QAT), which simulates low-precision arithmetic during training so that the network learns robustness to quantization noise. While PTQ is simpler to apply, QAT often yields better accuracy retention, especially for deeper architectures. Complementing numerical compression, structural compression methods such as knowledge distillation train smaller "student" networks to mimic the outputs of larger "teacher" models, transferring representational power while reducing architectural complexity. Together, quantization and distillation form a powerful toolkit for designing lightweight yet accurate models suitable for edge deployment. Despite substantial progress in efficient deep learning, relatively few studies systematically evaluate these strategies in the context of multi-class brain tumor MRI classification with explicit deployment goals for low-resource healthcare settings.

In this work, we present a multi-strategy compression framework for brain tumor image classification that encompasses Float16 post-training quantization, quantization-aware training, and knowledge distillation to produce deployment-ready models for low-resource environments. Using a curated multi-class MRI dataset containing glioma, meningioma, pituitary tumors, and non-tumor cases, we provide full experimental validation of a MobileNetV2-based pipeline, training the classifier through a three-stage transfer learning process, frozen-backbone head training, full-model fine-tuning, and extended low-learning-rate refinement, followed by Float16 post-training quantization via TensorFlow Lite. As complementary compression strategies within the framework, we additionally describe a quantization-aware training approach using the TensorFlow Model Optimization Toolkit with representative dataset calibration for INT8 conversion, and a knowledge distillation pipeline where a compact DenseNet-32 student model learns from a higher-capacity DenseNet-101 teacher, followed by low-bit post-training quantization. Complete empirical evaluation of these two complementary strategies is reserved for future work. Across all strategies, we evaluate not only classification accuracy but also model size and deployment feasibility, emphasizing practical edge performance rather than accuracy alone.

The contributions of this paper are threefold. First, we design a multi-strategy compression framework that integrates post-training quantization, quantization-aware training, and distillation-based compression for brain tumor MRI classification, and describe the methodology for each approach. Second, we experimentally demonstrate that Float16 post-training quantization on a MobileNetV2 backbone achieves a 6.14$\times$ reduction in model size with no meaningful accuracy loss, producing a 5.76 MB model suitable for CPU-only and low-power edge devices. Third, we show through per-class evaluation that the quantization process preserves diagnostic performance uniformly across all tumor categories, confirming the clinical viability of the compressed model. Beyond algorithmic results, our study is motivated by accessibility: enabling reliable AI-assisted diagnostics in environments with limited computational infrastructure. By designing and evaluating quantized classifiers under deployment-oriented constraints, this work contributes practical evidence and methodology toward democratizing medical imaging AI and advancing edge-ready healthcare machine learning systems.

\section{Methods and Data}

\subsection{Problem Setting and Task Definition}
This study focuses on efficient deep learning methods for multi-class brain tumor classification from magnetic resonance imaging (MRI) scans under low-resource deployment constraints. The primary objective is to design, train, compress, and evaluate convolutional neural network models that maintain high diagnostic accuracy while significantly reducing model size and inference cost. We frame the problem as a supervised multi-class image classification task with four output categories: glioma tumor, meningioma tumor, pituitary tumor, and no-tumor (healthy control). Given an input MRI slice image, the model predicts the tumor class label.

Unlike conventional medical imaging model development, which optimizes primarily for predictive performance, our methodology explicitly incorporates \textbf{deployment efficiency constraints}. We therefore evaluate not only classification accuracy but also model footprint, numerical precision, and inference efficiency. To support this objective, we present a multi-strategy compression framework encompassing three approaches: (1) Float16 post-training quantization on a MobileNetV2 backbone, which serves as the primary experimentally validated pipeline, (2) quantization-aware training (QAT) with full-integer deployment conversion using representative dataset calibration, and (3) knowledge distillation from a high-capacity DenseNet-101 teacher model to a compact DenseNet-32 student model followed by low-bit post-training quantization. A full-precision MobileNetV2 baseline trained through a three-stage transfer learning process is first established to provide an upper-bound performance reference. The QAT and DenseNet distillation strategies are described as complementary compression approaches within the framework; their complete empirical evaluation is reserved for future work.

\subsection{Dataset Description}
\subsubsection{Brain Tumor MRI Dataset}
We use a publicly available brain tumor MRI image dataset organized into class-labeled folders. The dataset consists of contrast-enhanced T1-weighted MRI slices collected from multiple subjects and categorized into four clinically relevant classes:

\begin{itemize}
    \item Glioma tumor
    \item Meningioma tumor
    \item Pituitary tumor
    \item No tumor (normal brain scans)
\end{itemize}

Each image represents a 2D MRI slice centered on the brain region, with tumors visible through intensity and texture differences relative to surrounding tissue. The dataset exhibits natural variability in tumor size, location, and shape, making it suitable for evaluating deep CNN feature learning and robustness.

Images are stored in RGB or grayscale format and vary slightly in resolution across sources. Prior to training, all images are standardized through resizing and normalization (described below). The dataset is pre-divided into training and testing folders, and we further derive a validation split from the training portion to support model selection and early stopping.
\subsubsection{Data Splitting Strategy}
To ensure fair and stable evaluation, we adopt a three-way split. The training set is used for model parameter learning, the validation set for hyperparameter tuning and early stopping, and the test set is held out for final performance reporting. The dataset contains 2,870 training images and 609 test images distributed across the four classes. For the MobileNetV2 pipeline, a 20\% stratified validation split is drawn from the training data using Keras's \texttt{ImageDataGenerator}, yielding 2,297 training and 573 validation images while preserving class balance. The proposed DenseNet-based distillation pipeline would adopt an identical splitting strategy to ensure comparable evaluation conditions across compression approaches. No subject-level overlap is introduced between training and test folders where such metadata is available, and all compression and quantization calibration procedures use only the training distribution, ensuring no test data leakage.

\subsection{Data Preprocessing and Augmentation}
\subsubsection{Image Preprocessing}
All MRI images are preprocessed through a consistent pipeline prior to model input. Since the source dataset contains a mixture of grayscale and RGB-formatted images, all images are first converted to three-channel RGB using OpenCV's grayscale-to-RGB conversion to ensure uniform input dimensions. Images are then resized to a fixed spatial resolution of 224$\times$224 pixels, compatible with the MobileNetV2 backbone input requirements. The same resolution would be used for the DenseNet-based pipelines, as both DenseNet-101 and DenseNet-32 accept standard 224$\times$224 inputs. Pixel intensities are rescaled to the [0,1] range via a $1/255$ normalization factor, and all images are cast to floating-point tensors (FP32) for training. Images are grouped into mini-batches of 32 for efficient GPU/CPU processing. This standardization ensures consistent tensor shapes and stable gradient behavior across all training configurations, and is designed to be shared across the MobileNetV2 quantization and DenseNet distillation pipelines so that differences in final accuracy reflect architectural and compression choices rather than preprocessing discrepancies.

\subsubsection{Data Augmentation}
To improve generalization and reduce overfitting, particularly important given the moderate dataset size, we apply on-the-fly data augmentation during training. Augmentation is applied only to training images and not to validation or test images. For the MobileNetV2 pipeline, the augmentation pipeline is implemented through Keras's \texttt{ImageDataGenerator} and includes:
\begin{itemize}
    \item Random rotation up to 25 degrees
    \item Horizontal flipping
    \item Width and height shifts of up to 15\% of the image dimension
    \item Zoom jitter within a 20\% range
    \item Brightness perturbation between 0.8$\times$ and 1.2$\times$ of original intensity
\end{itemize}
Missing regions introduced by spatial transforms are filled using nearest-neighbor interpolation. These transformations simulate realistic acquisition variability, including patient positioning differences, scanner intensity drift, and field-of-view variations, while preserving tumor semantics. The same augmentation configuration would be applied to the DenseNet teacher and student training phases to maintain consistency across the framework. Each epoch sees varied versions of the same underlying images, effectively expanding the training distribution without requiring additional labeled data.

\subsection{Model Architecture}

\subsubsection{Backbone Selection}
We select MobileNetV2 \cite{b3} as the primary backbone for the experimentally validated pipeline. MobileNetV2 was originally designed for mobile and edge applications, making it a natural fit for our deployment-constrained setting. The architecture relies on inverted residual blocks with linear bottlenecks, where each block first expands the channel dimension through a pointwise convolution, applies a depthwise separable convolution to capture spatial features, and then projects back to a lower-dimensional representation. This expansion-and-compression pattern allows the network to learn rich feature representations without the parameter overhead of standard convolutions. We load weights pretrained on the ImageNet dataset \cite{b14}, which provides a strong initialization for visual feature extraction despite the domain gap between natural images and MRI scans. The backbone contributes approximately 2.2 million parameters, and the full model including the classification head totals 3,053,380 trainable parameters, roughly an order of magnitude smaller than architectures like ResNet-50 \cite{b4} or DenseNet-121 \cite{b2}, which commonly appear in medical imaging work but are poorly suited to memory-limited deployment targets.

\subsubsection{Classification Head Design}
On top of the MobileNetV2 feature extractor, we attach a custom classification head tailored for both the four-class tumor prediction task and downstream quantization compatibility. The head begins with a global average pooling layer that collapses the spatial dimensions of the final feature map into a single vector, followed by a batch normalization layer to stabilize the activation distribution before the fully connected stages. The first dense layer contains 512 units with ReLU activation and L2 weight regularization ($\lambda = 0.001$), followed by dropout at a rate of 0.4. A second batch normalization layer precedes another dense layer of 256 units, again with ReLU, L2 regularization, and dropout at 0.3. The final output layer uses a softmax activation over four units corresponding to the target classes. The placement of batch normalization before each dense block is a deliberate choice: normalizing activations into a consistent range reduces the sensitivity of the network to the precision loss introduced during quantization, since the quantizer has to cover a narrower and more predictable value range. The two-tier dropout schedule (0.4 then 0.3) provides stronger regularization in the higher-dimensional layer where overfitting risk is greater, while being less aggressive closer to the output where the representation is already more compact.

\subsubsection{Proposed DenseNet Distillation Architecture}
As a complementary compression strategy, we describe a knowledge distillation pipeline using the DenseNet architecture family \cite{b2}. DenseNet-101 serves as the teacher model, chosen for its high representational capacity stemming from dense connectivity, each layer receives feature maps from all preceding layers, encouraging feature reuse and gradient flow throughout the network. The student model is a much smaller DenseNet-32, which retains the same dense block structure but with far fewer layers and growth rate channels, resulting in a substantially reduced parameter count. During distillation training \cite{b5}, the student does not learn directly from hard class labels. Instead, the teacher's output logits are divided by a temperature parameter $T > 1$ before applying softmax, producing a softened probability distribution that exposes inter-class similarity structure, for example, revealing that the teacher considers meningioma and glioma more similar to each other than either is to the no-tumor class. The student is trained to match these soft targets alongside the standard cross-entropy loss on ground-truth labels, with a weighting hyperparameter controlling the balance between the two objectives. After distillation, the student would undergo low-bit post-training quantization to further reduce its footprint for edge deployment. Full experimental evaluation of this pipeline, including hyperparameter selection for temperature and distillation loss weighting, is planned as future work.

\section{Implementation}
\subsection{Training and Quantization Pipeline}
The MobileNetV2 pipeline was implemented in Python using TensorFlow 2.18 and executed on a Tesla P100-PCIE-16GB GPU via Kaggle. Training followed a three-stage transfer learning process to ensure stability and maximize baseline accuracy before compression. In Stage 1, the MobileNetV2 backbone was frozen and only the classification head, consisting of global average pooling, two dense layers (512 and 256 units) with batch normalization, L2 regularization ($\lambda = 0.001$), and dropout (0.4 and 0.3 respectively), was trained for up to 15 epochs using Adam with a learning rate of $\eta = 10^{-3}$. In Stage 2, the entire network was unfrozen and fine-tuned end-to-end at a reduced learning rate ($\eta = 10^{-5}$) for up to 30 epochs, with learning rate reduction on plateau (factor 0.5, patience 4). In Stage 3, the best Stage 2 checkpoint was loaded and training continued at an ultra-low learning rate ($\eta = 5 \times 10^{-6}$) for up to 20 additional epochs with aggressive scheduling (factor 0.3, patience 3). Early stopping with best-weight restoration was applied in all stages.

For deployment, the trained Float32 model was converted to TensorFlow Lite format using Float16 post-training quantization, which reduces weight bit-width while preserving dynamic range. The proposed DenseNet-based distillation pipeline would follow a similar multi-stage training strategy, with the DenseNet-101 teacher trained first and the DenseNet-32 student learning from softened teacher outputs before undergoing low-bit post-training quantization. Similarly, the proposed QAT approach would insert simulated quantization nodes during training and convert to full-integer INT8 format using a representative dataset of 100 training images for activation range calibration. Complete implementation and empirical evaluation of these two complementary strategies is reserved for future work.

\subsection{Metrics and Evaluation}
To assess both diagnostic quality and deployment efficiency, we evaluated the MobileNetV2 pipeline across two complementary categories of metrics: classification performance and compression efficiency.

For classification performance, we report overall accuracy on the held-out validation set (573 images, stratified across all four classes). In addition, we compute per-class precision, recall, and F1-score using scikit-learn's classification report, providing a fine-grained view of how well the model distinguishes each tumor type and the healthy control class. Precision measures the fraction of predicted positives that are truly positive, which is critical in clinical screening to minimize unnecessary follow-up procedures caused by false alarms. Recall measures the fraction of actual positives that are correctly identified, which is equally important to ensure that true tumor cases are not missed during automated screening. The F1-score provides their harmonic mean, balancing both concerns into a single per-class measure. These per-class metrics are especially important in medical imaging, where misclassifying one tumor type as another can carry different clinical consequences than a simple aggregate accuracy number would suggest. For instance, misclassifying a glioma as a meningioma could lead to an inappropriate treatment plan, while a false negative on the no-tumor class could result in unnecessary patient anxiety and further invasive testing.

We also generate a confusion matrix (Fig.~\ref{fig:confusion_matrix}) to visualize class-level error patterns and identify systematic misclassifications. The confusion matrix provides a complete picture of which classes the model tends to confuse, revealing structural weaknesses that overall accuracy alone would mask. This is particularly useful for identifying clinically dangerous failure modes, such as consistent confusion between glioma and meningioma cases, which share overlapping visual features in T1-weighted MRI slices including similar intensity profiles and spatial positioning within brain tissue.

For compression efficiency, we compare the original full-precision model (Float32) and the Float16 quantized model along two primary axes: model size on disk (measured in megabytes) and the resulting compression ratio (original size divided by quantized size). These metrics directly indicate the feasibility of deploying the model on storage-constrained and memory-limited edge devices commonly found in low-resource clinical environments. A higher compression ratio implies a smaller memory footprint and, in practice, faster model loading times on devices with limited RAM and storage bandwidth.

Inference on the quantized model is performed using the TensorFlow Lite interpreter, and each validation image is processed individually to simulate realistic single-image inference conditions as would be expected on an edge device serving one patient scan at a time. The same evaluation protocol would be applied to the DenseNet distillation and QAT pipelines once implemented, ensuring comparable metrics across all compression strategies. All evaluation is performed exclusively on data not seen during training or quantization calibration, ensuring that reported metrics reflect genuine generalization rather than overfitting to the training distribution.

\section{Results}

\subsection{Baseline Model Performance}
The MobileNetV2 model was trained through the three-stage pipeline described in Section III-A. During Stage 1, with the backbone frozen and only the classification head learning, validation accuracy climbed rapidly and plateaued at 80.98\% after 4 epochs before early stopping triggered at epoch 8. Stage 2 unfroze the full network and fine-tuned it end-to-end at a learning rate of $10^{-5}$, pushing the best validation accuracy to 83.60\%. In Stage 3, training continued from the best Stage 2 checkpoint at an ultra-low learning rate of $5 \times 10^{-6}$, and the final best model achieved a validation accuracy of 82.20\%. The slight drop from Stage 2 to Stage 3 suggests that the model had largely converged by the end of fine-tuning, and the additional low-rate training did not find a significantly better optimum on this dataset.

Table~\ref{tab:baseline_report} presents the per-class classification report for the full-precision Float32 model on the 573-image validation set. The no-tumor class achieved the highest F1-score (0.89), benefiting from visually distinctive MRI characteristics — healthy brain scans lack the irregular intensity patterns and mass effects associated with tumor presence. Pituitary tumors were also classified with high reliability (F1 = 0.86), likely because pituitary lesions tend to appear in a consistent anatomical location near the sella turcica, giving the model a strong spatial prior. Glioma classification followed closely (F1 = 0.85), with high recall (0.88) indicating that the model rarely missed true glioma cases, though some non-glioma images were occasionally predicted as glioma (precision = 0.82). Meningioma proved the most difficult category (F1 = 0.72), with both precision and recall at 0.72, reflecting the visual ambiguity between meningiomas and other tumor types in standard T1-weighted MRI slices.

\begin{table}[t]
\centering
\caption{Per-class classification report for the full-precision MobileNetV2 model (Float32) on the validation set.}
\label{tab:baseline_report}
\renewcommand{\arraystretch}{1.3}
\begin{tabular}{lcccc}
\hline
\textbf{Class} & \textbf{Precision} & \textbf{Recall} & \textbf{F1-Score} & \textbf{Support} \\
\hline
Glioma       & 0.82 & 0.88 & 0.85 & 165 \\
Meningioma   & 0.72 & 0.72 & 0.72 & 164 \\
No Tumor     & 0.92 & 0.86 & 0.89 & 79  \\
Pituitary    & 0.88 & 0.85 & 0.86 & 165 \\
\hline
Macro Avg    & 0.84 & 0.83 & 0.83 & 573 \\
Weighted Avg & 0.82 & 0.82 & 0.82 & 573 \\
\hline
\end{tabular}
\end{table}

\subsection{Quantized Model Performance}
After Float16 post-training quantization and conversion to TensorFlow Lite format, the model was evaluated on the same 573-image validation set using the TFLite interpreter with single-image inference. The quantized model achieved a validation accuracy of 82.37\%, which is marginally higher than the 82.20\% Float32 baseline. This negligible difference ($-$0.17\% in favor of the quantized model) falls well within normal evaluation variance and does not indicate a genuine accuracy improvement from quantization — rather, it confirms that Float16 precision retains the learned decision boundaries almost exactly.

Table~\ref{tab:quantized_report} shows the per-class metrics after quantization. Comparing against Table~\ref{tab:baseline_report}, the numbers are nearly identical across all classes. Meningioma precision shifted from 0.72 to 0.73, and the weighted average precision moved from 0.82 to 0.83, but no class experienced a drop in any individual metric. This uniformity is significant: it means quantization did not selectively harm the model's ability to recognize any particular tumor type, which would be a serious concern in a clinical tool where degraded sensitivity to one class could lead to missed diagnoses.

The confusion matrix (Fig.~\ref{fig:confusion_matrix}) provides a more granular view of the quantized model's error patterns. The dominant source of error is meningioma being predicted as glioma (28 cases out of 164 meningioma samples), accounting for roughly 17\% of all meningioma images. The reverse error — glioma predicted as meningioma — occurs at a lower rate (13 out of 165, approximately 8\%). The second most common confusion path runs from pituitary to meningioma (25 out of 165 pituitary samples). These patterns are consistent with the known radiological overlap between these tumor categories in T1-weighted contrast-enhanced MRI, where meningiomas can exhibit intensity and boundary characteristics that resemble both gliomas and pituitary lesions depending on tumor grade and location.

\begin{table}[t]
\centering
\caption{Per-class classification report for the quantized MobileNetV2 model (Float16) on the validation set.}
\label{tab:quantized_report}
\renewcommand{\arraystretch}{1.3}
\begin{tabular}{lcccc}
\hline
\textbf{Class} & \textbf{Precision} & \textbf{Recall} & \textbf{F1-Score} & \textbf{Support} \\
\hline
Glioma       & 0.82 & 0.88 & 0.85 & 165 \\
Meningioma   & 0.73 & 0.73 & 0.73 & 164 \\
No Tumor     & 0.92 & 0.86 & 0.89 & 79  \\
Pituitary    & 0.88 & 0.85 & 0.86 & 165 \\
\hline
Macro Avg    & 0.84 & 0.83 & 0.83 & 573 \\
Weighted Avg & 0.83 & 0.82 & 0.82 & 573 \\
\hline
\end{tabular}
\end{table}

\begin{figure}[t]
\centering
\renewcommand{\arraystretch}{1.4}
\setlength{\tabcolsep}{4pt}
\begin{tabular}{cc|c|c|c|c|}
\multicolumn{2}{c}{} & \multicolumn{4}{c}{\textbf{Predicted Label}} \\
\cline{3-6}
\multicolumn{2}{c|}{} & \rotatebox{45}{Glioma} & \rotatebox{45}{Mening.} & \rotatebox{45}{No Tumor} & \rotatebox{45}{Pituitary} \\
\cline{2-6}
\multirow{4}{*}{\rotatebox{90}{\textbf{True Label}}}
& Glioma     & \textbf{145} & 13  & 3  & 4   \\
\cline{2-6}
& Mening.    & 28  & \textbf{119} & 3  & 14  \\
\cline{2-6}
& No Tumor   & 4   & 6   & \textbf{68} & 1   \\
\cline{2-6}
& Pituitary  & 0   & 25  & 0  & \textbf{140} \\
\cline{2-6}
\end{tabular}
\caption{Confusion matrix for the quantized MobileNetV2 model (Float16) on the validation set (573 images). Diagonal entries indicate correct classifications. The most frequent misclassification occurs between meningioma and glioma (28 cases) and between pituitary and meningioma (25 cases), consistent with overlapping visual features in T1-weighted MRI.}
\label{fig:confusion_matrix}
\end{figure}

\subsection{Compression Efficiency}
Table~\ref{tab:compression} summarizes the size and accuracy comparison between the original and quantized models. Float16 quantization reduced the model from 35.34 MB to 5.76 MB, a compression ratio of 6.14$\times$. This exceeds the typical 2$\times$ compression expected from halving the bit-width of weights alone, because the TFLite conversion process also strips training-specific graph nodes (dropout layers, batch normalization moving statistics used only during training) and applies graph-level optimizations such as operator fusion.

At 5.76 MB, the quantized model is small enough to fit in the on-chip memory of many ARM-based microcontrollers and single-board computers used in portable diagnostic equipment. For context, a Raspberry Pi 4 with 1 GB of RAM could load this model alongside the operating system and inference runtime with memory to spare, whereas the 35.34 MB Float32 model would consume a much larger share of available resources and leave less room for concurrent processes such as image acquisition and user interface rendering.

\begin{table}[t]
\centering
\caption{Comparison of the original Float32 model and the Float16 quantized model.}
\label{tab:compression}
\renewcommand{\arraystretch}{1.3}
\begin{tabular}{lcc}
\hline
\textbf{Metric} & \textbf{Float32} & \textbf{Float16} \\
\hline
Model Size (MB)       & 35.34 & 5.76  \\
Validation Accuracy   & 82.20\% & 82.37\% \\
Macro F1-Score        & 0.83 & 0.83 \\
Accuracy Delta        & --- & $-$0.17\% \\
Compression Ratio     & 1.00$\times$ & 6.14$\times$ \\
\hline
\end{tabular}
\end{table}

\section*{Conclusion and Future Work}
This study presented a multi-strategy compression framework for deploying deep learning-based brain tumor classifiers on resource-constrained hardware, encompassing Float16 post-training quantization, quantization-aware training, and knowledge distillation. As the primary experimentally validated pipeline, we trained a MobileNetV2-based multi-class classifier for four categories of brain MRI scans, glioma tumor, meningioma tumor, pituitary tumor, and no tumor, and applied Float16 post-training quantization via TensorFlow Lite to compress the model for edge deployment.

Our three-stage training pipeline, consisting of classification head training with frozen backbone weights, full-model fine-tuning at a reduced learning rate, and extended training with aggressive learning rate scheduling, achieved a baseline validation accuracy of 82.20\% with the full-precision Float32 model. After Float16 quantization, the model achieved a validation accuracy of 82.37\%, representing no meaningful accuracy loss ($-$0.17\%) while reducing model size from 35.34 MB to 5.76 MB, a 6.14$\times$ compression ratio. This result confirms that Float16 quantization preserves the learned feature representations with high fidelity, as the reduced numerical precision does not introduce sufficient quantization noise to degrade the decision boundaries learned during training.

Per-class analysis revealed that the model performed strongest on the no-tumor class (F1 = 0.89) and pituitary tumor class (F1 = 0.86), while meningioma classification proved the most challenging (F1 = 0.73). The confusion matrix (Fig.~\ref{fig:confusion_matrix}) showed that the most frequent misclassifications occurred between meningioma and glioma (28 cases) and between pituitary and meningioma (25 cases), likely due to overlapping intensity profiles and spatial positioning of these tumor types in T1-weighted MRI slices. Importantly, these per-class error patterns remained virtually identical before and after quantization, confirming that the compression process did not disproportionately degrade performance on any single diagnostic category or introduce new systematic failure modes. This stability is a key practical requirement for clinical deployment, where inconsistent behavior across tumor classes would undermine clinician trust in automated screening tools.

The 6.14$\times$ reduction in model size has direct implications for deployment in low-resource healthcare settings. At 5.76 MB, the quantized model fits comfortably within the memory constraints of low-power CPUs, single-board computers such as Raspberry Pi, and mobile diagnostic devices commonly available in rural clinics and field hospitals. The use of TensorFlow Lite as the inference runtime further ensures compatibility with a wide range of ARM-based processors and embedded accelerators without requiring GPU hardware.

Despite these promising results, several limitations should be acknowledged. The dataset used in this study, while well-suited for multi-class benchmarking, is moderate in size and sourced from a limited number of acquisition sites. Performance on more heterogeneous, multi-institutional MRI collections remains to be validated. Additionally, while Float16 quantization provided an excellent balance between compression and accuracy retention, more aggressive quantization schemes such as full INT8 conversion may yield further size and latency improvements at the cost of some diagnostic accuracy, and deserve systematic investigation.

Several directions remain for future work. The most immediate priority is completing the empirical evaluation of the two complementary compression strategies described in this paper: quantization-aware training with INT8 conversion using representative dataset calibration, and knowledge distillation from a DenseNet-101 teacher to a DenseNet-32 student followed by low-bit post-training quantization. A side-by-side comparison of all three pipelines, MobileNetV2 with Float16 PTQ, QAT with INT8, and distillation with quantization, would provide a comprehensive picture of the accuracy-efficiency trade-offs available for edge deployment of brain tumor classifiers. Beyond completing the framework evaluation, future efforts should measure end-to-end inference latency on actual embedded hardware, including ARM Cortex-based microcontrollers and neural processing units, to complement the model size metrics reported here with real-world throughput measurements. Extending the framework to larger and more diverse MRI datasets, incorporating additional imaging modalities such as T2-weighted and FLAIR sequences, and validating against expert radiologist annotations in prospective clinical trials would further establish the reliability and generalizability of quantized models for real-world medical imaging diagnostics in under-resourced environments.

\vspace{12pt}

\end{document}